# AUTOMATIC APPLICATION LEVEL SET APPROACH IN DETECTION CALCIFICATIONS IN MAMMOGRAPHIC IMAGE

Atef Boujelben[1], Hedi Tmar[2], Jameleddine Mnif[1], Mohamed Abid[2]

[1] Numeric Archiving & Medical Imaging, National School of Medicine of Sfax, Tunisia
`Atef.Boujelben@fss.rnu.tn, jameleddine.mnif@rns.tn`

[2] CES-Computer, Electronic And Smart engineering systems design Laboratory, National School of Engineers of Sfax: Tunisia
`hedi.tmar@ceslab.org, mohamed.abid@enis.rnu.tn`

## ABSTRACT

Breast cancer is considered as one of a major health problem that constitutes the strongest cause behind mortality among women in the world. So, in this decade, breast cancer is the second most common type of cancer, in term of appearance frequency, and the fifth most common cause of cancer related death. In order to reduce the workload on radiologists, a variety of CAD systems; Computer-Aided Diagnosis (CADi) and Computer-Aided Detection (CADe) have been proposed. In this paper, we interested on CADe tool to help radiologist to detect cancer. The proposed CADe is based on a three-step work flow; namely, detection, analysis and classification. This paper deals with the problem of automatic detection of Region Of Interest (ROI) based on Level Set approach depended on edge and region criteria. This approach gives good visual information from the radiologist. After that, the features extraction using textures characteristics and the vector classification using Multilayer Perception (MLP) and k-Nearest Neighbours (KNN) are adopted to distinguish different ACR (American College of Radiology) classification. Moreover, we use the Digital Database for Screening Mammography (DDSM) for experiments and these results in term of accuracy varied between 60 % and 70% are acceptable and must be ameliorated to aid radiologist.

## 1. INTRODUCTION

Breast cancer whose region is difficult to be visually detected is a major cause of death among women [1]. So, the quality of radiologist judgment of whether the suspected region is normal, benign or malignant will not be guaranteed. So far, screening mammography has been the best available radiological technique for an early detection of breast cancer [2]. However, because of the large number of mammograms to be analyzed, radiologists can make false detections. Thus, there are new solutions of automatic detection pertaining to the problems of analysis that can be explored. In this context, Computer Aided Diagnosis (CADi) and Computer Aided Detection (CADe) are two systems that can solve these problems [3]. In fact, in medical imaging, particularly System based on CADi and CADe are important in terms of cancer diagnosis and detection quality. Particularly, in breast cancer detection, region is difficult to be detected. In fact, the quality of breast income the differentiation between region (benign or malign) and normal is complicated. We interest in this paper in CADe Systems to minimize the load of the radiologist. Especially, we interest on; firstly, visual morphologic and distribution characterisation of calcifications and secondly automatic detection of ROI (Region Of Interest) composed of all distribution of calcifications. After that, we characterise ROI by using statistical Method of Gray Level Co-occurrence Matrix (GLCM) to have decision. In this context, the texture analysis is used to distinct the ACR (American College of Radiology) classifications.

In this paper, we include detection of ROI approach in the process of mammograms diagnosis. The main purpose of this work is the elaboration of a CADe to reach an automatic identification of ROI and contribute to a better quality of analysis. Firstly, we show why and how to adapt Level Set-based approach in the case of detection. Secondly, we study the performance of texture features in a mammogram diagnosis process.

The remainder of this paper is organized as follows. Section 2 describes the state of the art in the detection methods. Then, section 3 presents method of image-processing, automatic initialization and an adaptation of Level Set approach in segmentation particularly in case of breast cancer detection. Next, section 4 presents the adopted method for analysis basing in texture features extraction. Afterwards, section 5 presents the results, in ACR classifications, obtained by the proposed scheme. Lastly, section 6 gives some concluding remarks and draws some future perspectives.

## 2. STATE OF THE ART:

The identification of breast region is important to improve the analysis process. So, micro-calcification and macro-calcifications appear in mammograms with different shape characteristics and distribution. Thus, detecting the region can give an idea about the nature of diagnosis. However, in the past several years there has been tremendous evolution in mammography process. In this context, two approaches are used in the literature: automatic detection and region segmentation. Concerning detection, Torrent et al. [4] presents a comparison of three algorithms for segmenting fatty and dense tissue in mammographic images. The first algorithm is a multiple thresholding algorithm based on the excess entropy, the second one is based on the Fuzzy C-Means clustering algorithm, and the third one is based on a statistical analysis of the breast. In addition, Rangayyan et al. [5] presents a schema for the analysis of linear directional components in images by using a multi-resolution representation based on Gabor wavelets. A dictionary of Gabor filters with varying tuning frequency and orientation, which is designed to reduce the redundancy in the wavelet–based representation, is applied to the given image. The filter responses for different scales and orientation are analyzed by using the Karhunen–Loeve (KL) transform and Otsu's method of thresholding. The application of KL method is adopted to select the principal components of the filter responses, preserving only the most relevant directional elements appearing at all scales. The first N principal components, threshold are used to reconstruct the magnitude and the phase of directional components of the image by using Otsu's method. The proposed scheme is applied to the analysis of asymmetry between left and right mammograms. In this context, another method based on multiresolution approach to the computer aided detection of clustered micro-calcifications in digitized mammograms based on Gabor elementary functions is illustrated in [6]. To extract micro-calcifications characteristics a bank of Gabor functions with varying spatial extent and tuned to different spatial frequencies is used. Firstly, results show that most micro-calcifications, isolated or clustered, are detected and secondly the classification is illustrated by an Artificial Neural Network with supervised learning. On the other hand, Thangavel et al. [7] present an Ant Colony Optimization (ACO) and Genetic Algorithm (GA) for the identification of suspicious regions in mammograms. The proposed method uses the asymmetry principle (bilateral subtraction): strong structural asymmetries between the corresponding regions in the left and right breasts are taken as evidence for the possible presence of micro-calcifications in that region. Bilateral subtraction is achieved in two steps: first, mammograms images are enhanced using median filter, then pectoral muscle region is removed and the border of the mammogram is detected for both left and right images from the binary image. Further GA is applied to enhance the detected border. The figure of merit is calculated to identify whether the detected border is exact or not. So, the nipple position is identified for both left and right images using GA and ACO, and their performance is studied. Second, using the border points and nipple position as the reference of mammogram images are aligned and subtracted to extract the suspicious region. In the context of detection ROI, Schiabel et al. [8] proposed a methodology based on the Watershed transformation, which is combined with two other procedures: histogram equalization, working as pre-processing for enhancing images contrast, and a labelling procedure intended to reduce noise. However, the method based on fuzzy region growing is illustrated in [9]. The procedure starts with a seed pixel, and uses a fuzzy membership function based on statistical measures of the growing region. The results of testing with several mammograms indicate that the method can provide boundaries of tumours close to those drawn by an expert radiologist. The regions obtained preserve the transition information present around the tumour boundaries. Statistical measures computed from the resulting regions have shown the potential to classify masses and tumours as benign or malignant. In this context, fuzzy region detection is also adopted in [21]. But, Jadhav et al. [10] used statistical feature extraction method by using a sliding window analysis for detecting circumscribed masses in mammograms. This procedure is implemented by taking into account the multi-scale statistical properties of the breast tissue, and succeeds in finding the exact tumour position by performing the mammographic analysis using first few moments of each window. We have demonstrated that fast implementation in both feature extraction and neural classification module can be achieved. However, Tweed et al. [11] present an algorithm that selects ROI containing a tumour, based on the combination of a texture and histogram analysis. The first analysis compares texture features extracted from different regions in an image to the same features extracted from known timorous regions. The second analysis detects the ROI with two thresholds computed from the histograms of known timorous masks. So, the texture analysis is also used in [12][29]. Nevertheless, a system processes for the mammograms in several steps is adopted in [13]: first, the original picture is filtered with contrast shape which is sensitive to micro-calcifications. Then, authors enhance the mammogram contrast by using wavelet-based sharpening algorithm. Afterwards, present to radiologist for visual analysis, such a contrast-enhanced mammogram with suggested positions of micro-calcifications cluster. However, a multi-resolution representation of the original mammogram is obtained using a linear phase non-separable 2-D wavelet transform which is adopted in [14]. This is chosen for two reasons: first, it does not introduce phase distortions in the decomposed images and second, no bias is introduced in the horizontal and vertical directions as a separable transform would. Authors used coefficients of the analysis low pass filter. Then a set of features are extracted at each resolution for every pixel. After that, detection is performed from the coarsest resolution to the resolution using binary tree classifiers. This top-down approach requires less computation by starting with the least amount of data and propagating detection results to finer resolutions. In addition, wavelet coefficients describe the local geometry of an image in terms of scale and orientation apart from being flexible and robust with respect to image resolution and quality [15]. In this context, wavelet coefficients are also adopted in [16]. In addition, Marti et al. [12] propose a supervised method for the segmentation of masses in mammographic images. Based on the active region approach, an energy function which integrates texture, contour and shape information is defined. Then, pixels are aggregated or eliminated to the region by optimizing this function allowing the obtention of an accurate segmentation. The algorithm starts with a selected pixel inside the mass, which has been manually selected by an expert radiologist.

Recently, explicit and implicit methods of deformable model are used in different applications [17]. In this context, for breast cancer detection, Ferrari et al. [18] used a traditional active deformable contour model (Snake) to limit the breast in the image. To injure the problem of initialization, they used an adaptative thresholding. For another need, particularly the elimination of the pectoral muscle, Boucher et al. [19] used the snake and Ball et al. [20] used the Narrow Band level set methodology with an adaptative segmentation threshold controlled by a border complexity term.

An overview of the literature shows that many other methods of segmentation and identification are used to detect ROI [30]. In this paper, we propose method based on Level Set approach, an implicit method of deformable model, which includes edge and region proprieties. So, in the Level Set approach, two major problems are usually discussed in the bibliographies: initialization and evolution function which is the point of interest in the next section.

## 3. MICROCALCIFICATIONS DETECTION:

In this section, we show how to adopt region and edge criteria in Level Set approach detection to characterize shape of calcifications. However, before the automatic initialization which we can start Level Set method, we extract breast in the mammographic image to eliminate noise causes with DDSM (Digital Database for Screening Mammography) images.

### 3.1 Breast extraction:

To identify the breast region, we used successive steps given by the block diagram shown at figure 1. In this section a short description of each step is presented.

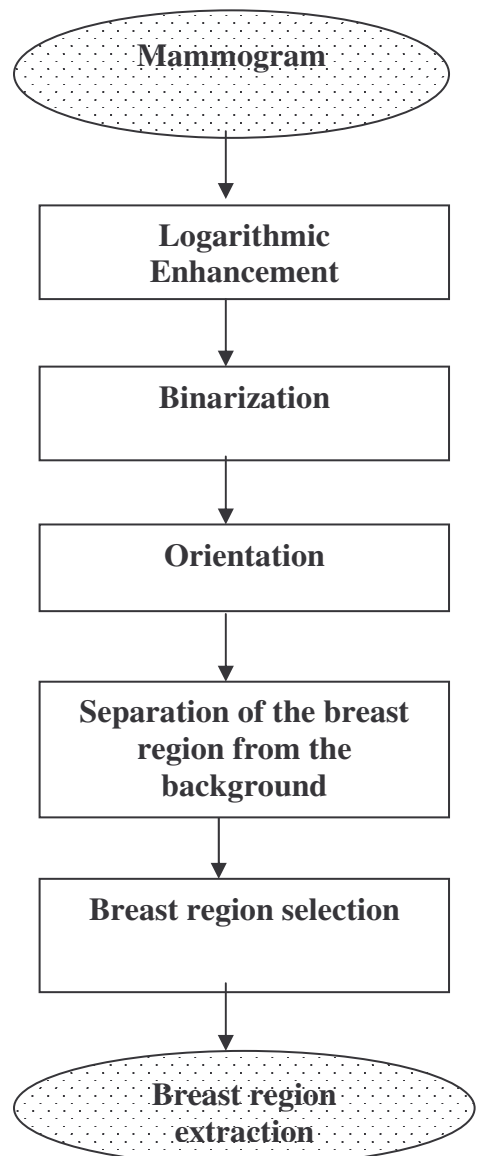

Figure 1. Block diagram of breast region extraction.

In the following subsections we will illustrate each stage included in the pre-processing procedure.

### 3.1.1 Logarithmic enhancement

In mammography, the application of the logarithmic transform to the whole image significantly enhances the contrast of the regions near the breast boundary in mammograms, which are characterized by low density and poor definition of details [32][33]. In our approach the logarithmic transform of pixel I(x,y) has the form:

$$G(x, y) = (c * \log(I(x, y) - s_{min} + 1)) + 1 \quad (1)$$

Where c is a normalization factor following by:

$$c = \frac{-2}{\log(s_{max} - s_{min} + 1)} \quad (2)$$

- Smin and Smax are the minimum and the maximum pixel values of the input image.
- G(x, y) is the transformed pixel.

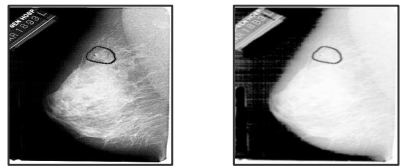

(a)  (b)

Figure 2. Logarithmic enhancement result: (a) original image; (b) enhanced image.

The objective of application logarithmic transformation to the mammogram image is to determine an approximate breast contour as close as possible to the true breast boundary. We show in figure 2 an example of Logarithmic enhancement. The next step is the binarization which is the point of interest in the next sub-section.

### 3.1.2 Binarization

In this step, we use an automatic thresholding method to obtain a binarization of the enhanced image. In this context, we used three thresholding methods for applying on each image in order to choose the best: the maximum-entropy principle [22], Otsu's method [23], and a method based on the maximum correlation criterion [33]. After application of each method, we found that the best result was given by Otsu's method. In figure 3, we present an example of the result of binarization given by each method mentioned above.

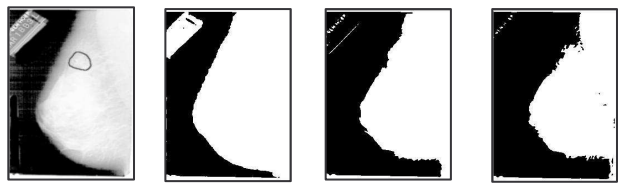

(a)  (b)  (c)  (d)

Figure 3. Binarization result: (a) Enhanced image; (b) Otsu' method; (c) Maximum-entropy principle method; (d) Maximum correlation criterion method.

### 3.1.3 Orientation

This step permits to identify the orientation of the breast region. For that, we divide the image into two equal parts (figure 4) and we calculate the number of pixels of each part: if number of white pixels is big in left part, the direction of breast region is from left to right (respectively, if the number of white pixels is big in right part the direction of breast region is from right to left).

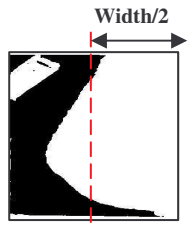

Figure 4. Image divided in two equal parts.

### 3.1.4. Separation of the breast region from the background

In most images, the breast region is connected with tape artefact. For this reason, this step is used to separate the breast region from the background (tape artefact and labels). In first phase (figure 5(a)), we search the two points (A and B) that coincide with the breast region: we take two parts from the mammogram image; a part at the top and another part at the bottom of the image (each part have as height the 1/12 of the image height). We travel the image vertically (from top to bottom); we start from the beginning of the image and we check if all pixels are white. We stop when we encounter the first black pixel, that is the searched point (noted A). Similarly to the research of the point B, except that we cross the image inversely (bottom to top).

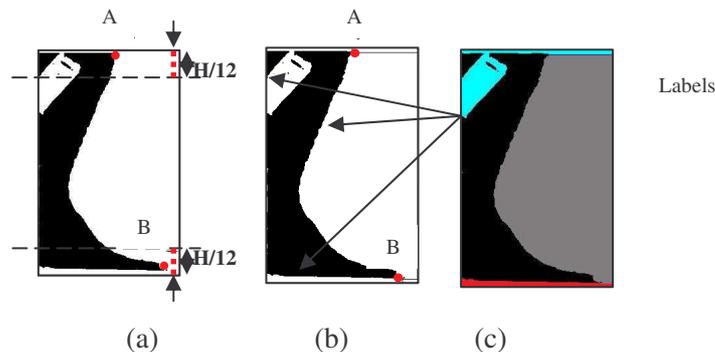

(a) (b) (c)

Figure 5. Separation of the breast region from the background: (a) Identification of top and bottom points; (b) Drawing of two lines for separation; (c) Using of the Connected Component Labelling algorithm.

In a second phase, after the localisation of two points, we draw two lines: the first point is between the beginning of the image (from left to right or from right to left) and the point A while the second is between the beginning of the image and the point B (figure 5(b)). Finally, we use the connected component labelling algorithm [25] to divide the binary image into different labels (figure 5c)).

### 3.1.5 Breast region selection

Looking at the image generated by the last step, we note that breast region has the largest area. For this raison, we use the area criteria to select the label that represents the breast region and to eliminate the unlikely labels (tape artefacts and high intensity labels). Finally, to obtain the effective breast region, the result of this step was multiplied with the original mammogram. Figure 6 shows, with details, the results of the various steps in the breast extraction stage.

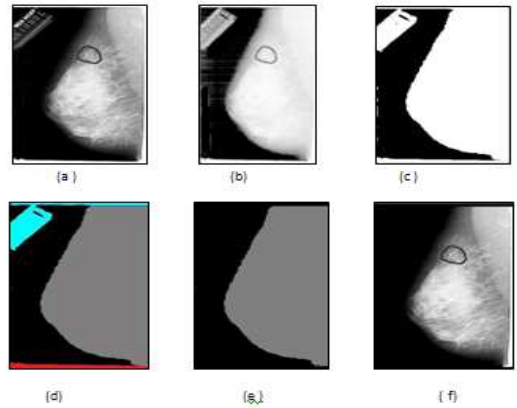

Figure 6. Breast region extraction: (a) Original image; (b) Logarithmic enhancement; (c) Binarization with Otsu's thresholding; (d) Separation of the breast region; (e) Selection of the largest label; (f) Final result of breast extraction.

### 3.2 Adaptation Level set Approach in detection of ROI:

In level set approach, there are two aspects of research which are the initialization and the function of evolution. So, Level Set is a method which studied evolution of the curve and surfaces [26]. The points defining this interface will move towards the normal at a speed F according to the following equation:

$$\frac{\partial c(t)}{\partial t} = F \vec{N} \quad (3)$$

$\vec{N}$ : Normal with the curve
F: Speed term depending on the curve

The parametric curve C(t) is recovered by the detection of the level zero of, respectively, the function F evolves and moves according to:

$$\frac{\partial c(t)}{\partial t} = F \|\nabla \phi\| \quad (4)$$

The evolution of this function depends on an initial curve $\phi 0$. In this case two aspect of research: initialization and the function F. Generally [6][27], speed F depends on three terms: firstly on the local curve in each point (pondered with $\varepsilon$), term dependent on the image (pondered with $\beta$) and a constant term (pondered with $v$). The evolution of interface is given by the following equation:

$$\frac{\partial c(t)}{\partial t} = \varepsilon * g(I) \|\nabla \phi\| - \beta * \nabla g(I) \|\nabla \phi\| + v * g(I) \|\nabla \phi\| \quad (5)$$

Where: I is the point (i,j) of image matrix
$\varepsilon, \beta, v \in [0, 1]$
$$g(I) = \frac{1}{1 + \|\nabla I\|} \quad (6)$$

To minimize the temporal complexity of this equation, we adopt the Narrow Band and Fast Marching method in implementation. Narrow band consists of computing Level Set on evolution from contour for early inside and outside near the Level Set zero [28]. We use this approach for two reasons; firstly, to optimize time computation efficiency for numerical calculus Level set method. Secondly, in general, regions in breast are difficult to be detected. In fact, we should focus locally near to the zero Level Set and its neighbouring Level Set because the local contour has more information significance than distant ones to get visual calcifications information to radiologist. However, to accelerate the convergence of Level Set approach, we adopt a monotonically advancing front based on Fast Marching approach [28]. Its idea is if T(x,y) is the time at which the curve crosses the point (x,y) so the surface T(x,y) then satisfies the equation:

$$\|\nabla T\| F = 1 \tag{7}$$

Where: $F = \frac{1}{e^{-\alpha \nabla I}}$ (8)

This equation allows a good implementation of deformable contours. Indeed, the changes of topology are managed automatically. Thus, if the image contains several objects, contour is divided during its evolution to include each object separately. Contour can also become deformed to be adjusted with complex forms, which cannot do explicit active contours (Snakes). Another positive point is that this method does not depend on initialization.

However, in the case of textured images, criteria of gradient (edge proprieties) in which depends this equation (uniformity inter-region) affected an over-segmentation. So, the presence of textures in a mammographic image generates bad results because the small areas are privileged. But one can have resorts to a measurement of containing area in order to improve the quality of calcifications detection.

In this context, region propriety is adopted firstly with the notion of image and secondly with the notion of propagation (addition of a fourth term). The evolution of interface is actually, which has ameliorated equation 5, given by the following equation:

$$\frac{\partial \phi}{\partial t} = \varepsilon^* g(I) \|\nabla \phi\| . div \frac{\nabla \phi}{|\nabla \phi|} - \beta^* (\nabla g(I) \|\nabla \phi\| + \frac{Moy(I)}{Max(I)}) + \upsilon^* g(I) \|\nabla \phi\| - \theta^* SkewCentredNormal(I) \tag{9}$$

Where: $\theta \in [0,1]$
Max(I)=maximum of gray-level in image
Moy(I)=average of gray-level 3*3 centred in (x,y)

$$SkewCentredNormal(I) = \frac{SkewCentred}{Max(Skewness)} \tag{10}$$

The SkewCentred corresponds to the moment around the average. It measures the deviation of the distribution of the gray-level compared to a symmetrical distribution.

$$SkewCentred = \frac{1}{9} \sum_x \sum_y (I(x,y) - Moy(x,y)) \tag{11}$$

For a deviation to raised values, the Skewness-Centred is positive; whereas for a deviation towards low values, it is negative. In figure 7, we represent one malignant case in type ACR5 (more detail with ACR classification is showed in section 4): so we have respectively originally image in DDSM database, breast extracted, application Level Set approach in new image and ROI extracted. This figure shows the result of detection ROI, using initialization-points with maximum Gray-Level. Firstly, this result shows radiologist visually points of view in diagnostic results. In fact, using Level Set approach with an edge and a region criterion can give an idea with morphologic and distribution of calcifications. Secondly, ROI is detected automatically; a sliding windows limited with two points (P1(X-minimum, Y-maximum), P2(X-maximum, Y-minimum)) of all small regions detected.

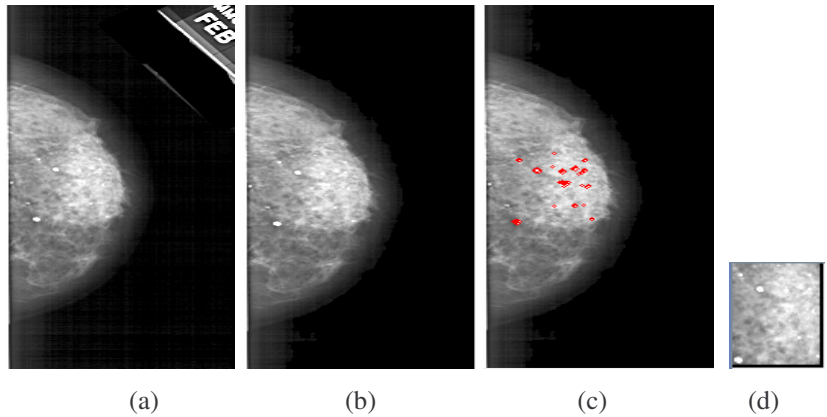

(a) (b) (c) (d)

Figure 7: detection of ROI: (a) Original image in DDSM database, (b) Breast detection, (c) Level Set application with maximum gray Level initialization, (d) ROI detected.

After the ROI detection, the extraction of features is adopted in ROI: this is the point of interest that we will focus on, in the next section of this paper.

## 4. ANALYSIS OF ROI

In the ROI detection, we use an adaptation of a Level Set Approach with an edge and a region criterion. In this section, the features extraction is illustrated on any ROI. So, ROI included all calcifications which its have different distributions and shape characteristics.

### 4.1 Calcifications characteristics:

Calcifications have different shapes and distributions in the mammograms. So, we can distinct between macro-calcifications (Figures 8) and micro-calcifications (Figures 9).

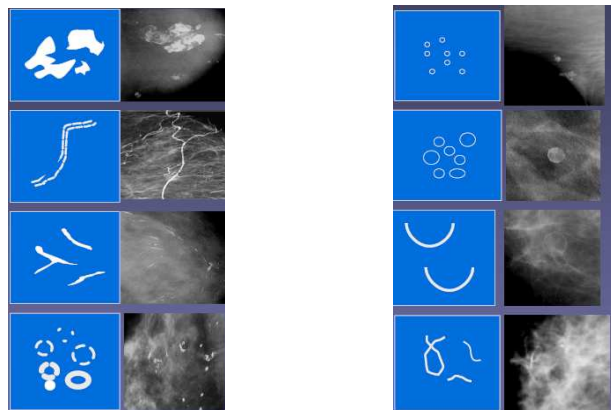

Figure 8: Macro-calcifications types

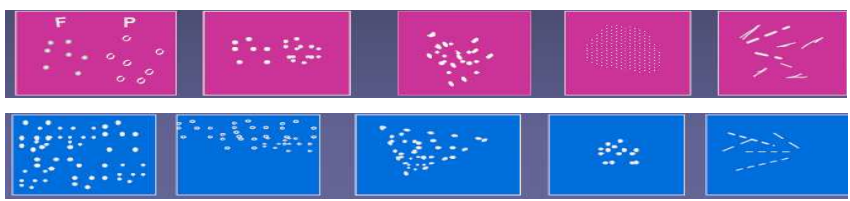

(a) Morphologic

(b) Distributions

Figure 9: Micro-calcifications morphologic and distributions

In the context of calcification classification, there are five standards of assessment categories from Breast Imaging Reporting and Data System (BI-RADS):

- ACR1 (Negative): without calcifications,
- ACR2 (Benign finding): macro-calcifications without opacity or annular, lunar, vascular annular, formed a deposit of micro-calcifications,
- ACR3 (Probably benign): micro-calcifications round punctiform regular or pulverulent, very few, isolated round clusters,
- ACR4 (Suspicious abnormality): many micro-calcifications pulverulent irregular, very few polymorphic,
- ACR5 (Highly suggestive of malignancy): micro-calcifications vermicular, irregular, grouped or micro-calcifications with topography different or micro-calcifications evolutionary.

For characterization of morphological and the distribution of calcifications, we introduce a method based on texture analysis to distinct ROIs.

### 4.2 Texture analysis

To determine morphological and distribution qualities, we adopt the texture method. In this context, a texture feature extraction, which is frequently cited in the literature, is based on the use of CGLM. Co-occurrence matrix is a second-order statistical measure of image variation. In this subsection, we detail the feature of co-occurrence approach. We represent our analysis by statistical texture. From this approach, we extract six characteristics which are defined as follows:

$$moy = \frac{1}{N} * \sum_x \sum_y p(x, y) \qquad (12)$$

where: p(x,y) denotes the gray-level in the co-occurrences matrix.

$$Variance = \sum_x \sum_y (x - moy)^2 \, p(x, y) \qquad (13)$$
$$Energy = \sum_x \sum_y p(x, y)^2 \qquad (14)$$
$$Contrast = \sum_x \sum_y (x - y)^2 \, p(x, y) \qquad (15)$$
$$Entropy = -\sum_x \sum_y p(x, y) \log p(x, y) \qquad (16)$$
$$Homogeneity = \sum_x \sum_y \frac{1}{1 + (x - y)^2} \, p(x, y) \qquad (17)$$

The algorithm evaluates the properties of the region in mammographic image. We investigate the performance of feature in texture from GLCM in diagnosis by using four orientations 0, 45, 90, 135. From each one, we inspect six features (then we take the average of one feature of the four orientations).

In the next section, we will show the performance of the textural vector in analyzing ROI in terms of diagnosis relevance by using kNN and MLP classifiers.

### 5. RESULTS AND DISCUSSION:

The terminology which is used to determine the performance of a CADe System is accuracy: percentage of correctly classified cases. Because of the variation in the types of breast cancer, a large number of cases can reduce the dependency of analysis techniques versus image sets. The performance of an algorithm is affected by the characteristics of a database like the digitization techniques which are namely pixel size, subtlety of cases, choice of training/testing subsets, etc.

### 5.1 The DDSM dataset

The establishment of the DDSM allows the possibility of the common training and testing Dataset. The DDSM is the largest publicly available database of mammographic data. It contains approximately 2620 screening mammography cases. From the total number of mammographic images included in the DDSM database, we use 500 images decomposed with 100 in witch ACR. To make a good evaluation, we use the remaining 300 images which are divided into 60 in witch ACR.

To classify the ROI, which included calcifications distribution, detected with Level Set approach using DDSM dataset with textural-vector illustrated in the least section, one will use two classifiers which are kNN (K=7) and MLP. In the next subsection, we illustrate the results of accuracy with an analysis method.

### 5.2 Experimental results: Performance in terms of diagnosis quality

The basic classification is based on two methods of classification KNN (K=7) and MLP as shown in Table 1. It represents the results from texture analysis in ROI detected.

Table 1: Results from analysis based on texture description vector

|      | KNN    | MLP    |
|------|--------|--------|
| ACR1 | 66,66% | 65%    |
| ACR2 | 70%    | 63,33  |
| ACR3 | 66,66% | 61,66  |
| ACR4 | 60%    | 60%    |
| ACR5 | 61,66  | 63,33% |

Table 1 shows the diagnostic performance from the automatic detection method using ACR classification method for diagnosis. So, in the first steps we have best visual information from the radiologist. But, in the second steps we have weak percentage of accuracy. So, the accuracy varied between 60% and 70% using KNN classifier, and, 60% and 65% using MLP classifier. This method must be ameliorated witch used as the second diagnosis reference with radiologist. However, in comparing these results with other related work, we notice the majority used two classifications categories namely malignant and benign ones.

So, in terms with accuracy the percentage of these results are not the best result compared to local works [34][35]. In this context, in [34] the result is about 94% in boundary information; and in [35] the result of accuracy is between 90% and 92% using extended Radial Distance Measure method. But, firstly we used two types of classifications namely malignant or benign and secondly, in such work, we used DDSM database but the ROI is selected from the image by fixing a rectangular box around the suspicious lesion area and the classical method of segmentation based on Sobel filter and thresholding. In the other hands, other related works are sited. In fact, Alvarenga et al. [36] obtained 88% of sensitivity (percentage of pathological ROIs which is correctly classified) and 90% of specificity (percentage of non-pathological ROIs which is correctly classified). In their experiments, they used a local images dataset and Linear Discriminant Analysis (LDA) method for classification. Additionally, Rangayyan et al. [37] have used the LDA classifier and their result reaches 95% in terms of classification accuracy. Conversely, the result of Retico et al. [38] using a MLP classifier can reach 78.1% and 79.1% in sensitivity and specificity, respectively. In the others hand, using a SVM classifier, Chang et al. [39] obtained respectively 88.89% and 92.5% in terms of sensitivity and specificity.

Yet, the characterization of mammographic masses and tumours and their classification as being benign or malignant is difficult. But, this difficulty increase witch using ACR classifications. So that, we can assume the acceptable of results especially for the nature of cases in DDSM databases witch using cases composed with 4 images can have ACR different. But these results must be ameliorated to aid radiologist.

## 6. CONCLUSION

In this work, we attempted to improve the automatic detection ROI of in the process of breast cancer diagnosis. In this paper we introduced an adaptation of Level Set approach to detect ROI witch combining edge and region criteria. Firstly, we show a pre-processing step to eliminate noise in the DDSM images. Secondly, an automatic initialization starting with pixel having values of level of gray maximum. In fact, the point is located in the area (intra-area). Thirdly, Level Set is propagated and adapted locally inside the area and will be blocked in the edges to determine morphologic of calcifications. This result give a good visually information to radiologist. Finally, to determine the distribution of calcification, an automatic detection of ROI is adopted with showing small-windows included all calcifications. However, the detection of ROI is the first step of CADe system followed by the analysis steps, using texture approach, and classifications steps, using KNN and MLP classifiers. The results obtained are encouraging, but we can ameliorate classifications steps by using the advantages of Bayesian-Network which we can integrate private information of patients.

**Authors**

**Atef Boujelben** is a researcher at CES "Computer, Electronic and Smart engineering systems design Laboratory, In National School of Engineers of Sfax" and "Numeric Archiving & Medical Imaging, In National School of Medicine of Sfax". University of Sfax, Tunisia.
His current research interests are Computer Aided Decision, medical image processing, computer graphics and multimedia.

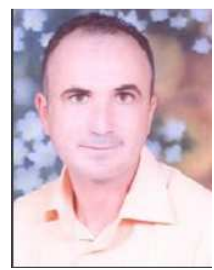